\documentclass[11pt, a4paper, logo, copyright]{googledeepmind}

\usepackage[authoryear, sort&compress, round]{natbib}
\usepackage{graphicx} 
\usepackage{subcaption} 
\usepackage{fancyhdr} 
\usepackage{lipsum} 
\usepackage{caption} 
\usepackage{color}

\usepackage{cleveref}
\usepackage{afterpage}

\usepackage{booktabs}

\usepackage{threeparttable}

\title{Evidence of Cognitive Deficits and Developmental Advances in Generative AI: A Clock Drawing Test Analysis}

\author[$\dagger$,1]{Isaac R. Galatzer-Levy}
\author[3]{Jed McGiffin}
\author[2]{David Munday}
\author[1]{Xin Liu}
\author[2]{Danny Karmon}
\author[2]{Ilia Labzovsky}
\author[2]{Rivka Moroshko}
\author[2]{Amir Zait}
\author[1]{Daniel McDuff}

\affil[$\dagger$]{Corresponding Author}
\affil[1]{Google Research}
\affil[2]{Google DeepMind}
\affil[3]{University of Washington}

\begin{abstract}
The rapid advancement of generative artificial intelligence (GenAI) has ignited significant interest in the cognitive capabilities of the underlying models. This surge in interest stems from the unprecedented ability of these models to perform tasks previously considered exclusive to human cognition, such as natural language understanding, creative writing, and code generation. Interrogating the performance of such models on standardized cognitive tasks can provide insights into how they directly compare to healthy and abnormal human performance. This study explores the performance of multiple recent GenAI models on the Clock Drawing Test (CDT), a widely used neuropsychological assessment for evaluating aspects of executive functioning such as visuospatial planning and organization. Our findings demonstrate that while models can produce clock-like representations, they exhibit fundamental deficits in the ability to reason and produce the correct time, consistent with mild to severe cognitive deficits using a standardized scoring scheme \citep{wechsler2009}.  Specifically, AI-generated drawings frequently demonstrate errors in numerical sequencing (omissions, repetitions, misordering), and numerical reasoning (clock time errors), and intrusions (addition of irrelevant or hallucinated information) while models consistently perform well in rendering features (numbers, hands, contour), together indicating crystallized knowledge but an inability to demonstrate visual-spatial reasoning. Of those evaluated, only GPT 4 Turbo and Gemini Pro 1.5 successfully produced the correct time and demonstrated scores consistent with healthy cognitive functioning (Weighted score = 4/4). A follow-up test in which multimodal models were asked to read a clock revealed that only one model (Sonnet 3.5) read the clock, correctly indicating that the deficits in drawing are due to an inability to comprehend, attend, or manipulate numeric concepts. The observed findings can reflect deficits in visual-spatial understanding, working memory, and/or calculation. Together, this research identifies consistent strengths in crystallized knowledge but deficits in reasoning capabilities. This underscores the value of benchmarks that facilitate comparison across human and machine performance. This translation allows us to understand the cognitive capabilities of artificial intelligence and to guide further development towards generalized artificial cognitive functions akin to those of neurological healthy humans.  
\end{abstract}

\begin{document}

\maketitle

\section{Introduction}

Generative artificial intelligence (GenAI) models possess the ability to perform tasks that resemble human cognitive functions, such as reasoning and information retrieval. This capacity stems from their training on massive data sets of text and code, enabling them to learn complex patterns and relationships. Unlike the biologically constrained architecture of the human brain, GenAI models operate on computational principles that are still being deciphered. Comparing their performance on cognitive tasks with human benchmarks allows us to probe these underlying mechanisms and elucidate the computational underpinnings of intelligence.  Although generative models can produce remarkably human-like text and images, their underlying cognitive architecture differs significantly from ours and remains poorly understood. Just as human cognitive abilities are rooted in neuroanatomy yet emerge as complex, measurable behaviors, understanding the cognitive capabilities of these models is essential.  Only then can we grasp their potential and limitations as simulations of human cognition and their capacity for everyday reasoning.

The Clock Drawing Test (CDT, \citep{agrell1998clock}) is a classic performance-based neuropsychological screening tool, offering a concise yet multifaceted evaluation of cognitive domains frequently compromised in neurodegenerative disease and other brain disorders. Its sensitivity to a range of cognitive functions makes it an ideal instrument for probing the capabilities of GenAI models. By examining their performance on this test, we can gain insights into their strengths and weaknesses in areas such as visuospatial processing, planning, and numerical reasoning. The deceptively simple task of drawing a clock face and setting the hands to a specified time, engages a complex interplay of cognitive processes, including visuospatial skills (the ability to accurately represent spatial relationships; \citep{agrell1998, fukui2009}, executive functioning, (ability to plan, sequence, organize, and reason; \citep{jones2021, dubois2008}, working memory, (critical for parsing  instructions and maintaining a mental image of a clock; \citep{bondi1996}, and sustained attention (ability to focus on the task and resist distractions; \citep{lezak1995}). Performance on the CDT is commonly used by psychologists and physicians to screen for cognitive dysfunction in the above domains, and is particularly sensitive to pathology associated with neurodegenerative disorders, including Alzheimer’s disease and other dementia subtypes 6–8. Given that such tasks can be translated for administration to generative models, such methods serve as useful benchmarks of cognition that translate across humans and machines.  

Error analysis in CDT extends beyond a simple numerical total score, offering insights into specific cognitive deficits and potential underlying neuroanatomical mechanisms involved. For example, number omissions may suggest attentional impairment or executive dysfunction (e.g., poor error monitoring), whereas sequencing errors in number placement or hand setting often reflect executive dysfunction and frontal lobe involvement \citep{lezak1995}. Errors in number generation, such as repeated or incorrect numbers, can indicate difficulties with numerical sequence comprehension, working memory limitations, or visuospatial organization deficits. A distorted clock contour, or otherwise visually fragmented clock lacking overall “gestalt” is indicative of visuospatial dysfunction and potential parietal lobe involvement \citep{fukui2009}. A misplaced center, where the hands of the clock converge, suggests difficulty with visuospatial planning and organization. Finally, intrusions, such as extraneous drawings or words, point to executive dysfunction and poor response inhibition, commonly seen in frontal lobe disorders and sometimes associated with psychosis \citep{jones2021, royall1999}.

Performance on the CDT is not itself diagnostic of pathology as multiple factors influence cognitive performance. Moreover, impaired performance on any individual subtest should be considered amongst data from a broader neuropsychological assessment  for questions relevant to diagnosis. Limitations notwithstanding, various CDT scoring systems are available (for a review, see: \citep{spenciere2017scoring}) that have been normed against the general population and provide performance brackets indicating likely levels of cognitive impairment. As such, population norming for tests refers to the percentile ranking of the cognitively healthy population that would attain a score at or below that score. This provides a reference for the chances that the individual being tested is cognitively healthy, or conversely, at risk requiring further focused evaluation. Referring to norms developed as part of the Brief Cognitive Status Exam (BCSE) of the Wechsler Memory Scale - Fourth Edition (\citep{wechsler2009} scores on the CDT (as elaborated in the methods section) can be used to rank examinees’ CDT total scores as ‘Very Low (corresponding to smaller than 2\% chance of being cognitively healthy),’ ‘Low (2-4\%),’ ‘Borderline (9\%),’ ‘Low Average,’ and ‘Average’. Scores in the very low range have a high probability of being considered abnormal while scores in the Low range have moderate probability of being abnormal. Scores in the Borderline range and higher have less evidence that scores indicate significant cognitive impairment. In this instance, the interpretation focuses on specific aspects of poor performance such as attention or inhibition \citep{wechsler2009}.  Importantly, while assessment of human performance can directly reflect known anatomical structures and functions, in the context of generative models, results only tell us about their current state of development and skill level. 

\section{Methods}

\subsection{Models}
In the current investigation, the CDT was administered to multiple generative models independently. This included large language models (LLMs) that are capable of language understanding and generation and multimodal models that are capable of image generation from language understanding. 

The language models we tested (as of July 2024) were Google’s Gemini Nano, Gemini Pro, Gemini Ultra and Gemini Advanced, OpenAI’s PT-3.5 Turbo, GPT-4 Turbo, and Anthropic's Claude 3 Opus. The multimodal models were Google Deepmind’s Image Gen 2, Stable Diffusion XL Base 1.0, Stable Diffusion 3 Medium, OpenAI’s GPT-4o. These represent a range of foundation models with different numbers of underlying parameters, training data and regimes, and architectures. We did not assess any models in the Llama family developed by Meta because licensing requires direct permission from Meta to use such models for research purposes.

\textit{Language only models}
\textbf{Google Gemini Family:} We utilized four models from Google's Gemini family: Gemini Nano, Gemini Pro, Gemini Ultra, and Gemini Advanced.  These models are designed for different computational constraints and offer varying levels of capability. However, Google has not publicly released details about the specific architectures, training data, or number of parameters for these models. 
\textbf{OpenAI GPT Family:} We incorporated two models from OpenAI's GPT family: GPT-3.5 Turbo and GPT-4 Turbo. GPT-3.5 Turbo, a text-only model. GPT-4 Turbo, is a larger more advanced model although OpenAI has not disclosed specifics regarding its architecture, training dataset size, or number of parameters.
\textbf{Anthropic:} Claude 3 Opus by Anthropic is also an open source large language model known for its focus on safety and helpfulness. 

\textit{Multimodal Models:}
\textbf{Google Gemini Family:} 
\textit{Google Deepmind ImageGen 2:} We employed Google Deepmind’s ImageGen 2, a text-to-image diffusion model. ImageGen 2 demonstrates significant advancements in generating high-quality images from text prompts.
\textit{Stability AI Stable Diffusion XL Base 1.0 \& Stable Diffusion 3 Medium:} We incorporated two models from Stability AI's Stable Diffusion suite: Stable Diffusion XL Base 1.0 and Stable Diffusion 3 Medium. These open-source latent text-to-image diffusion models are known for their ability to generate high-resolution images.
\textbf{OpenAI} 
\textit{GPT-4o:} We utilized GPT-4o, OpenAI's multimodal model capable of processing and generating both text and images. 

\subsection{Administration and Scoring}

All models were similar in that specific parameter counts and training approaches are not disclosed publically. Across multimodal and LLM models, the same prompt was administered to the model directly (once per model) with no previous prompting or fine tuning. The prompt for the CDT, as incorporated within the WMS-IV BCSE \citep{wechsler2009}, is to “draw the face of a clock, put in the numbers, and set the hands to 10 minutes after nine” \citep{wechsler2009}. For LLMs, additional instructions were provided to render code to produce the clock using SVG (text added to prompt: “Provide Scalable Vector Graphics (SVG) code to draw the image.”). 

The CDT involves three distinct components: drawing the clock face, putting in the numbers, and setting the hands. In the WMS-IV BCSE system, the resultant drawing is then scored on individual performance criteria resulting in  a total raw score ranging from 0 to 15. CDT raw scores are  then converted to Weighted Raw Score are scores that are bucketed into reliable ranges related to pathological presentations ranging from 0 (Very Poor) to 4 (Normal) based cognitive functioning \citep{wechsler2009}. Points are assigned in four categories including: 

\begin{itemize}
\item Numbers: Participants are scored on
\begin{itemize}
\item The order by which they place numbers on the clock.
\item The presence of numbers.
\item The location of the number placement.
\end{itemize}
\item Contour:
\begin{itemize}
\item Presence of a contour or circle representing the clock face.
\item Accommodation of the contour such that it is large enough to fit the clock numbers and hands inside.
\item Closure of the contour.
\item Symmetry of the shape.
\end{itemize}

\item Hands: 
\begin{itemize}
\item Presence of exactly two hands.
\item Connection of the hands to each other. \item Proportion of the hands such that one is noticeably shorter than the other.
\item Correct placement of the hand such that the short hand points to 11 and the long hand points to two. 
\end{itemize}

\item Center: 
\begin{itemize}
\item Presence of a center focal point of the clock. This center can be made by the connection of the two hands.
\item The location of the center in the middle of the contour. 
\end{itemize}
Additional information is recorded by the examiner that does not directly impact this scoring system but can provide diagnostic information such as intrusions in which hands of the clock are drawn as human-like hands or the clock is elaborated by additional images. 

As an additional step, models were asked to read a pre-generated image of a clock and report the correct time. This is a common follow-up test if performance on the CDT demonstrates any deficits. Asking the participant to read a clock allows for differential diagnosis between motor difficulties or deficits in neurological capabilities in production compared to difficulties in comprehension of underlying numeric concepts, and the ability to hold and manipulate numeric concepts.

\end{itemize}

\section{Results}
\begin{figure}[h!]
   \caption{Comparison of outputs from four different models on the CDT Test.}
        \label{fig:main}
        \vspace{0.5cm}
        \small 
    \centering
    \begin{minipage}{\textwidth}
        \centering
        \parbox{\textwidth}{
            \centering
            \hrule height 0.4pt 
            \vspace{0.2cm}
            \textbf{Multimodal Models (Image Generation; n = 4 models} 
            \vspace{0.2cm}
            \hrule height 0.4pt 
            \vspace{0.1cm}
        }
        \begin{subfigure}[t]{0.24\textwidth}
            \centering
            \includegraphics[width=\textwidth]{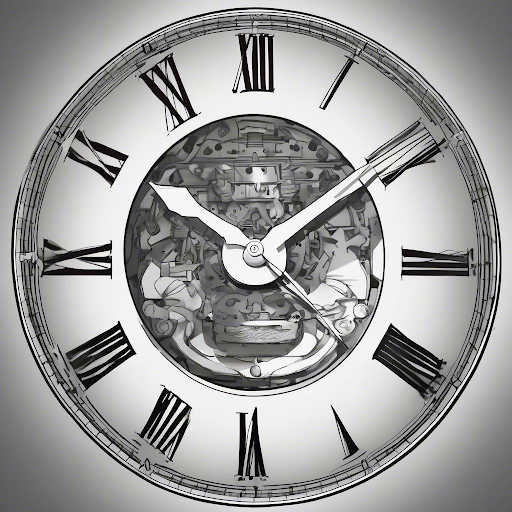} 
            \caption{Gemini Stable diffusion XL base 1.0; Weighted Score (3)}
            \label{fig:subimage-a}
        \end{subfigure}
        \hfill
        \begin{subfigure}[t]{0.24\textwidth}
            \centering
            \includegraphics[width=\textwidth]{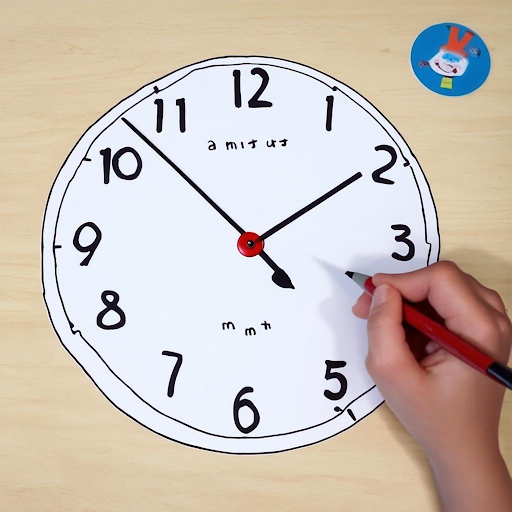} 
            \caption{Gemini Stable Diffusion 3 Medium; Weighted Score (2)}
            \label{fig:subimage-b}
        \end{subfigure}
        \hfill
        \begin{subfigure}[t]{0.24\textwidth}
            \centering
            \includegraphics[width=\textwidth]{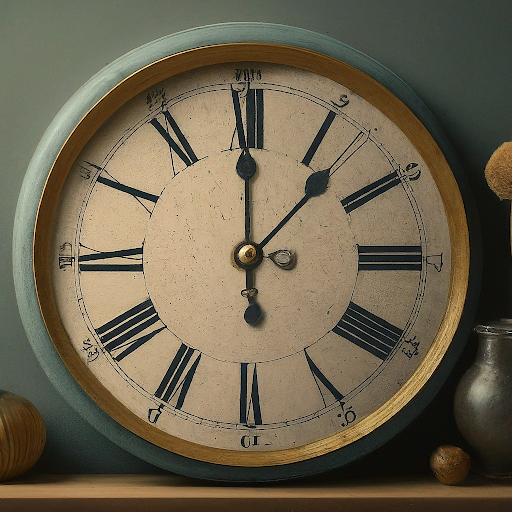} 
            \caption{Gemini Imagen 2; Weighted Score (3)}
            \label{fig:subimage-c}
        \end{subfigure}
        \hfill
        \begin{subfigure}[t]{0.24\textwidth}
            \centering
            \includegraphics[width=\textwidth]{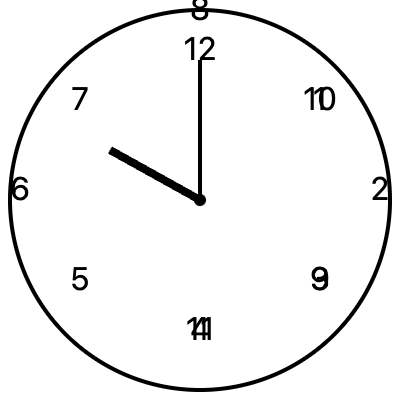} 
            \caption{GPT 4o; Weighted Score (4)}
            \label{fig:subimage-d}
        \end{subfigure}
        \parbox{\textwidth}{
            \centering
            \hrule height 0.4pt 
            \vspace{0.2cm}
            \textbf{Language Models (SVG Generated n = 9 models) } 
            \vspace{0.2cm}
            \hrule height 0.4pt 
            \vspace{0.1cm}
        }
                \begin{subfigure}[t]{0.24\textwidth}
            \centering
            \includegraphics[width=\textwidth]{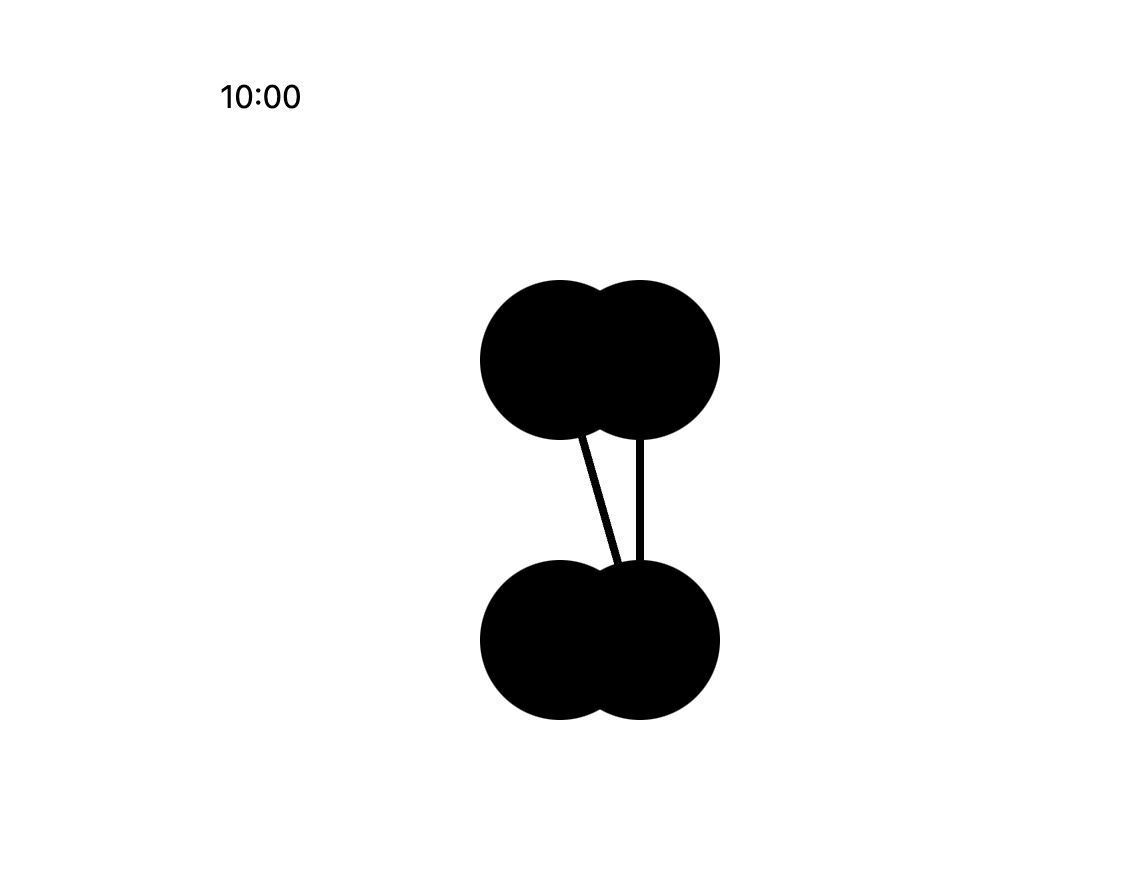} 
            \caption{Gemini Nano' ; Weighted Score (0)}
            \label{fig:subimage-ae}
        \end{subfigure}
        \hfill
        \begin{subfigure}[t]{0.24\textwidth}
            \centering
            \includegraphics[width=\textwidth]{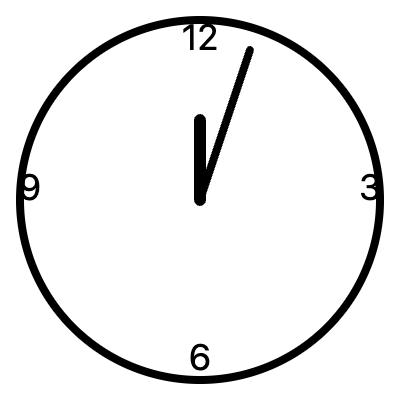} 
            \caption{Gemini Advanced; Weighted Score (3)}
            \label{fig:subimage-ab}
        \end{subfigure}
        \hfill
        \begin{subfigure}[t]{0.24\textwidth}
            \centering
            \includegraphics[width=\textwidth]{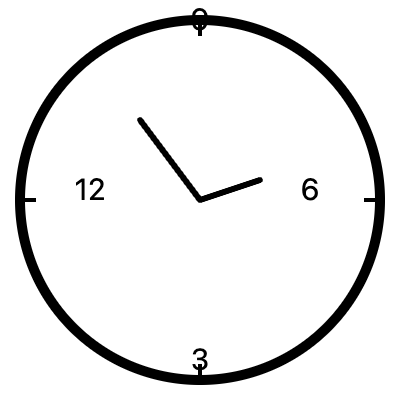} 
            \caption{Gemini Pro 1.0; ; Weighted Score (3)}
            \label{fig:subimage-bb}
        \end{subfigure}
        \hfill
        \begin{subfigure}[t]{0.24\textwidth}
            \centering
            \includegraphics[width=\textwidth]{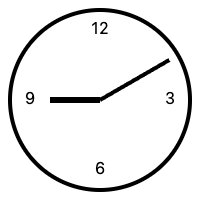} 
            \caption{Gemini Pro 1.5; Weighted Score (4)}
            \label{fig:subimage-cb}
        \end{subfigure}
        \hfill
        \begin{subfigure}[t]{0.24\textwidth}
            \centering
            \includegraphics[width=\textwidth]{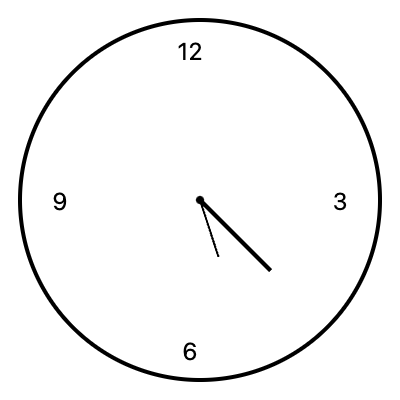} 
            \caption{GPT-3.5 Turbo; Weighted Score (4)}
            \label{fig:subimage-db}
        \end{subfigure}
                \begin{subfigure}[t]{0.24\textwidth}
            \centering
            \includegraphics[width=\textwidth]{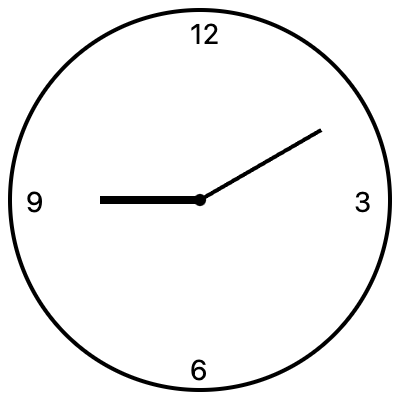} 
            \caption{GPT-4 Turbo; ; Weighted Score (4)}
            \label{fig:subimage-ac}
        \end{subfigure}
        \hfill
        \begin{subfigure}[t]{0.24\textwidth}
            \centering
            \includegraphics[width=\textwidth]{gpt_4o.png} 
            \caption{GPT 4o; Weighted Score (2)}
            \label{fig:subimage-bc}
        \end{subfigure}
        \hfill
        \begin{subfigure}[t]{0.24\textwidth}
            \centering
            \includegraphics[width=\textwidth]{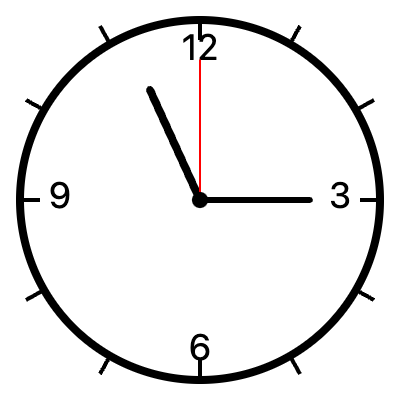} 
            \label{fig:subimage-cc}
        \end{subfigure}
        \hfill
        \begin{subfigure}[t]{0.24\textwidth}
            \centering
            \includegraphics[width=\textwidth]{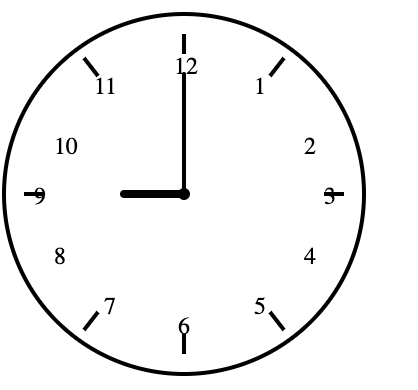} 
            \caption{Claude Sonnet 3.5 ; Weighted Score (4)}
            \label{fig:subimage-dc}
        \end{subfigure}
        \hfill
    \end{minipage}
\end{figure}

Variability in performance was observed across the 12 models which were each tested only once (Figure \ref{fig:main}, Table \ref{result_table}). Only the most advanced and highest parameter models (GPT 4 Turbo, Gemini Pro 1.5 demonstrated the capability to place both hands of the clock at the correct time. Larger parameter models were the only ones to demonstrate scores consistent with the population of individuals with healthy cognitive functioning [GPT 3.5 Turbo (Raw Score = 12; Weighted Score = 12; GPT 4 Turbo (Raw Score = 12; Weighted Score = 4; Gemini Pro 1.5 (Raw Score = 14; Weighted Score = 4)] while the smallest parameter model scored the poorest (Gemini Nano, Raw Score = 0; Weighted Score = 0). All models, with the exception of Gemini Nano which was unable to render a representation of an analogue clock or correct elements, demonstrated relatively minor errors in clock contour, number generation and sequencing, proportional hands and center alignment. Interestingly, GPT-4o was the only tested model able to perform both direct-visual and SVG-generated renderings, although these demonstrated a wide divergence in scores and neither was able to accurately render the correct time. Together these results indicate that, over a floor threshold of small parameterized models, current state-of-the-art multimodal and large language generative models are able to produce images grossly consistent with the concept of a clock and its constituent elements; however, only larger models, possibly with more advanced approaches to training, are currently able to produce clock drawings that also simulate the correct use of a clock to accurately represent human-level conceptualizations of time. 

Additionally, models either consistently omitted information with seven models failing to produce all 12 numbers. Conversely, multiple models demonstrated intrusions including novel shapes in the place of numbers and additional objects like human hands in the case of multimodal models, and additional numbers (GPT 4o language only). Additionally, Miro 4 and Miro 5 produced elaborate intrusions beyond simple changes to existing clock elements. 

Finally, when selected models (Gemini Pro 1.5, GPT 4 Turbo, GPT 4o, Claude Opus 3, Claude Sonnet 3.5) were shown a clock and asked to read the time, all failed with the exception of Claude Sonnet 3.5. Specifically, when presented with a clock that displays the time "5:45", two models demonstrated that they primarily attended to the long hand, misinterpreted its meaning, and ignored the short hand by stating that the time is "9:00"(Gemini Pro 1.5, GPT 4 Turbo). Other models (GPT-4o) confused the long and short hand to interpret the time as "9:30", as did Gemini XL reporting "9:25". Claude Opus 3 demonstrated a similar conceptual confusion about the long and short hands by interpreting the time as "4:50". Claude Sonnet 3.5 demonstrated the correct answer and demonstrating both the correct time and the underlying reasoning by stating "The clock in the image is showing 5:45.The hour hand is positioned between 5 and 6, but closer to 5, indicating it's 5 o'clock.The minute hand is pointing directly at 9, which represents 45 minutes past the hour.Therefore, the time shown on this analog clock is 5:45."

\begin{table}[ht]
\centering
\caption{Evaluation results for different models on the CDT test.}

\resizebox{\linewidth}{!}{
\begin{tabular}{p{2cm}p{0.7cm}p{0.7cm}p{0.7cm}p{0.7cm}p{0.7cm}p{0.7cm}p{0.7cm}p{0.7cm}p{0.7cm}p{0.7cm}p{0.7cm}p{0.7cm}p{0.7cm}p{0.7cm}p{0.7cm}c}
\toprule
Model/Item & \multicolumn{3}{c}{\textbf{1. Numbers}} & \multicolumn{4}{c}{\textbf{2.Contour}} & \multicolumn{4}{c}{\textbf{3.Hands}} & \multicolumn{2}{c}{\textbf{4.Center}} & \textbf{Total Raw} & \textbf{Weighted Raw} \\
 & Seq & Pres & Loc & Pres & Acc & Clos & Sym & Pres & Conn & Prop & Corr & Pres & Loc &  & \\ 

\midrule
\multicolumn{16}{l}{\textbf{Multimodal Models (Direct Image Generation)}} \\

Gemini Stable diffusion XL base 1.0 & 1 & 0 & 1 & 1 & 1 & 1 & 1 & 0 & 1 & 1 & 0 & 1 & 1 & 10 & 3 \\

Gemini Stable Diffusion 3 Medium & 1 & 0 & 1 & 1 & 1 & 0 & 1 & 0 & 1 & 1 & 0 & 1 & 1 &  11 & 3  \\

Gemini Image Gen 2 & 1 & 0 & 1 & 1 & 1 & 1 & 1 & 0 & 1 & 0 & 0 & 1 & 1 & 9 & 2 \\

GPT-4o & 2 & 1 & 1 & 1 & 1 & 1 & 1 & 1 & 1 & 1 & 0 & 1 & 1 & 13 & 4 \\
\midrule
\multicolumn{16}{l}{\textbf{Language Model (SVG Generated)}} \\

Gemini Nano & 0 & 0 & 0 & 0 & 0 & 0 & 0 & 0 & 0 & 0 & 0 & 0 & 0 & 0 & 0 \\

Gemini Pro 1.0 & 2 & 0 & 0 & 1 & 0 & 1 & 1 & 1 & 1 & 1 & 0 & 1 & 1 & 10 & 3 \\

Gemini Pro 1.5 & 2 & 0 & 1 & 1 & 1 & 1 & 1 & 1 & 1 & 2 & 1 & 1 & 1 & 14 & 4 \\

Gemini Adv. & 2 & 0 & 0 & 1 & 1 & 1 & 1 & 1 & 1 & 0 & 0 & 1 & 1 & 11 & 3 \\

GPT-3.5 Turbo & 2 & 0 & 1 & 1 & 1 & 1 & 1 & 1 & 1 & 0 & 0 & 1 & 1 & 12 & 4 \\

GPT-4 Turbo & 2 & 0 & 1 & 1 & 1 & 1 & 1 & 1 & 1 & 2 & 1 & 1 & 1 & 14 & 4 \\

GP-4o & 0 & 0 & 0 & 1 & 1 & 1 & 1 & 1 & 1 & 0 & 0 & 1 & 1 & 9 & 2 \\

Claude & 2 & 0 & 0 & 1 & 1 & 1 & 1 & 0 & 1 & 1 & 0 & 1 & 1 & 10 & 3 \\

\bottomrule

\end{tabular}
}
\label{result_table}
\caption{Clock drawing elements were evaluated according to the WMS-IV BCSE scoring criteria within the following four domains:  \textbf{ (1) Numbers}, whether these were displayed in correct Sequence (Seq; Range 0-2), including whether the model placed anchoring numbers first (12,3,6,9; optimal approach, 2 point score), Presence (Pres, range 0-1) indicating the presence of all 12 numbers in the image, Location (Loc 0-1) indicating numbers were correctly placed so as to be readable around the interior perimeter of the clock; \textbf{(2) Contour}, including whether appropriate contour is Present ( (Pres, range 0-1), whether the contour is large enough to fit all key elements, Accommodation (Acc, range 0-1), and Symmetry (Sym, range 0-1) whether the face is roughly symmetrical; \textbf{(3) Hands}, including whether two clock hands were Present  (Pres, range 0-1), whether two clock hands meet at a central focal point and are Connected (Conn, range 0-1), whether the clock hands are correctly proportioned (Prop, range 0-1) with exactly one short and one long hand, and whether the appropriate hands point to the correct time (Corr, range 0-1); \textbf{(4) Center}, including whether the clock has a clear central focal point that is present  (Pres, range 0-1), and Location (Loc, range 0-1), whether the center is roughly discernible in the middle of the clock. Scores are converted from a total raw score to a population weighted raw score to aid in interpretation. }
\end{table}


\section{Discussion}
Results indicate that generative AI models consistently demonstrate performance impairments (errors, omissions, and intrusions) on the Clock Drawing Test, a common screening tool sensitive to  abnormal cognitive functioning in humans. While all models successfully generated basic clock components like the contour, hands, and center, suggesting intact basic visuospatial processing and motor planning \citep{agrell1998}, significant deficits emerged in higher-order cognitive functions. The consistent misplacement of hands across all models, regardless of image or SVG generation, implies a shared difficulty in accurately representing spatial and temporal concepts, a cognitive domain reliant on intact executive functioning and working memory \citep{bondi1996, libon1993}. Other models exhibited errors mirroring human cognitive deficits, such as potential working memory limitations reflected in number omissions or sequencing errors \citep{bondi1996, libon1993}.This finding aligns with previous research demonstrating challenges in AI models' grasp of abstract concepts like time \citep{xu2023, dasgupta2022}. The superior performance of larger, more recent models suggests that increasing model scale and refining training methodologies contribute to advancements in higher-order cognitive abilities in GenAI. Of note, larger parameter more current models performed the best, both able to generate the concept of a clock and able to represent a clock functionally. This indicates that, though generative models are overall unreliable in tasks that involve reasoning, there is progress across generations indicating that generative models may soon reliably perform higher order cognitive tasks. 

Additional analysis of model’s ability to read a clock revealed that the deficits observed in on the CDT were not simply due to the inability to produce an image correctly. Indeed, the models demonstrated a deficit in the ability to understand the underlying numeric concepts of an analog clock and to manipulate them for their purposes. Some models simply reported an unrelated time while others demonstrated an inability to differentiate the hands of a clock to read the clock correctly. Only Claude Sonnet demonstrated the ability to read the clock correctly. While it is unknown why this model performed better, the output was accompanied by logical steps and their individual answers indicating that the model had accompanying prompts to guide its reasoning. This indicates that some internal mechanisms to break down and logically solve parts of the problem may be a successful strategy for generative models to perform higher order reasoning tasks such as this. This highlights the importance of incorporating explicit reasoning mechanisms into AI models and indicates that some internal mechanisms to break down and logically solve parts of the problem is a successful strategy for generative models to perform higher order reasoning tasks.

The presence of intrusions  in the form of extraneous images or numbers suggests potential weaknesses in response inhibition and attention, key components of executive function often associated with frontal lobe function in humans \citep{agrell1998, royall1999} regardless of method of producing an output. Clinically such a pattern of behavior can indicate either a lack of focused attention on the primary task or over-fixation of some small unimportant aspect at the expense of the primary goal which are both behaviors that are common in attentional disorders like Attention Deficit Hyperactive Disorder(ADHD) or early forms of dementia. Notably, while these intrusions resemble patterns observed in individuals with frontal lobe disorders, it's crucial to avoid direct comparisons given the inherent differences between AI models and the human brain.

It is important to acknowledge the limitations of this study. Primarily, the small sample size of models tested limits the generalizability of these findings to the broader landscape of generative AI. Second, we elected a single CDT administration for each AI model to stay consistent with the traditional testing format that underlies the population norming, although future replications of these results with iterative administrations per model are warranted to accurately assess the range and stability of error profiles associated with each model. Additionally, while the CDT is a sensitive cognitive screening tool, it provides a limited representation of human cognitive processes that are otherwise complex and multimodal. Future research should expand on these findings by incorporating a larger, more diverse set of models with iterative administrations, and utilizing a more comprehensive array of neuropsychological assessments to evaluate the strengths and weaknesses of AI across human cognitive domains. Finally, a significant limitation is the lack of publicly available information about the models themselves, limiting the ability to understand the relationship between performance and key parameters such as training data or the size of the model. While we can observe that smaller models perform more poorly and newer model perform better, we can not directly compare parameter count or other key features to cognitive ability. 

Importantly, while poor performance in humans reflects underlying biological deficits, limitations observed in generative models can only be interpreted as a snapshot of the current capabilities along a developmental trajectory. Indeed, we observe that as models advance in size and versioning, there is a developmental trend towards improved performance without any known specific training to solve the problem of clock drawing or recognition. In this context, poor performance can be interpreted as as limited development rather then an insurmountable deficit.  

Despite these limitations, this study offers valuable insights into the evolving capabilities and limitations of generative AI models. The observed patterns of performance on the CDT, a task sensitive to various cognitive domains, provide a unique lens through which to understand how these models process and integrate information as well as the targets for development that will improve general cognitive performance. As AI technology continues to advance, such investigations will be essential in guiding development, refining our understanding of AI's cognitive abilities, and ultimately bridging the gap between artificial and human intelligence. Importantly, the Clock Drawing Task indicates broad meta-cognitive capabilities and deficits. Further research should investigate specific substrates of cognition including executive functioning, working memory, visual reasoning, and crystallized knowledge, all of which can affect performance on this task.

\bibliography{main}

\begin{thebibliography}{13}
\providecommand{\natexlab}[1]{#1}
\providecommand{\url}[1]{\texttt{#1}}
\expandafter\ifx\csname urlstyle\endcsname\relax
  \providecommand{\doi}[1]{doi: #1}\else
  \providecommand{\doi}{doi: \begingroup \urlstyle{rm}\Url}\fi

\bibitem[Agrell and Dehlin(1998)]{agrell1998clock}
B.~Agrell and O.~Dehlin.
\newblock The clock-drawing test.
\newblock \emph{Age and ageing}, 27\penalty0 (3):\penalty0 399--404, 1998.

\bibitem[Agrell and Dehlin(2012)]{agrell1998}
B.~Agrell and O.~Dehlin.
\newblock The clock-drawing test.
\newblock \emph{Age Ageing}, 41\penalty0 (Suppl 3):\penalty0 iii 41--5, 2012.

\bibitem[Bondi et~al.(1996)Bondi, Salmon, and Kaszniak]{bondi1996}
M.~Bondi, D.~Salmon, and A.~Kaszniak.
\newblock The neuropsychology of dementia.
\newblock In \emph{Neuropsychological assessment of neuropsychiatric and
  neuromedical disorders}. 1996.

\bibitem[Dasgupta and et~al.(2022)]{dasgupta2022}
I.~Dasgupta and et~al.
\newblock Language models show human-like content effects on reasoning.
\newblock \emph{arXiv preprint arXiv:2207.07051}, 2022.

\bibitem[Dubois et~al.(2008)Dubois, Andrade, and Levy]{dubois2008}
B.~Dubois, K.~Andrade, and R.~Levy.
\newblock Executive dysfunction and neuropsychological testing.
\newblock In \emph{Handb. Clin. Neurol.}, volume~89, pages 35--52. 2008.

\bibitem[Fukui and Lee(2009)]{fukui2009}
T.~Fukui and E.~Lee.
\newblock Visuospatial function is a significant contributor to functional
  status in patients with alzheimer’s disease.
\newblock \emph{Am. J. Alzheimers. Dis. Other Demen.}, 24:\penalty0 313--321,
  2009.

\bibitem[Jones and Graff-Radford(2021)]{jones2021}
D.~T. Jones and J.~Graff-Radford.
\newblock Executive dysfunction and the prefrontal cortex.
\newblock \emph{Continuum}, 27:\penalty0 1586--1601, 2021.

\bibitem[Lezak(1995)]{lezak1995}
M.~D. Lezak.
\newblock \emph{Neuropsychological Assessment}.
\newblock Oxford University Press, New York, NY, 1995.

\bibitem[Libon et~al.(1993)Libon, Swenson, Barnoski, and Sands]{libon1993}
D.~J. Libon, R.~A. Swenson, E.~J. Barnoski, and L.~P. Sands.
\newblock Clock drawing as an assessment tool for dementia.
\newblock \emph{Arch. Clin. Neuropsychol.}, 8:\penalty0 405--415, 1993.

\bibitem[Royall et~al.(1999)Royall, Mulroy, Chiodo, and Polk]{royall1999}
D.~R. Royall, A.~R. Mulroy, L.~K. Chiodo, and M.~J. Polk.
\newblock Clock drawing is sensitive to executive control: a comparison of six
  methods.
\newblock \emph{J. Gerontol. B Psychol. Sci. Soc. Sci.}, 54:\penalty0 P328--33,
  1999.

\bibitem[Spenciere et~al.(2017)Spenciere, Alves, and
  Charchat-Fichman]{spenciere2017scoring}
B.~Spenciere, H.~Alves, and H.~Charchat-Fichman.
\newblock Scoring systems for the clock drawing test: A historical review.
\newblock \emph{Dementia \& neuropsychologia}, 11\penalty0 (1):\penalty0 6--14,
  2017.

\bibitem[Wechsler(2009)]{wechsler2009}
D.~Wechsler.
\newblock \emph{WMS‐IV Technical and Interpretive Manual}.
\newblock Pearson, 2009.

\bibitem[Xu et~al.(2023)Xu, Li, Vaezipoor, Sanner, and Khalil]{xu2023}
Y.~Xu, W.~Li, P.~Vaezipoor, S.~Sanner, and E.~B. Khalil.
\newblock Llms and the abstraction and reasoning corpus: Successes, failures,
  and the importance of object-based representations.
\newblock \emph{ArXiv abs/2305.18354}, 2023.

\end{thebibliography}

\end{document}